\pdfoutput=1

\documentclass[11pt]{article}

\usepackage[preprint]{acl}

\usepackage{times}
\usepackage{latexsym}
\usepackage{amsmath}
\usepackage{amssymb}
\usepackage{colortbl}
\usepackage{booktabs}
\usepackage{multirow}
\usepackage{appendix}

\usepackage[T1]{fontenc}

\usepackage[utf8]{inputenc}

\usepackage{microtype}

\usepackage{inconsolata}

\usepackage{graphicx}

\usepackage{hyperref}
\usepackage{url}
\usepackage{tikz}
\usepackage{pgfplots}
\pgfplotsset{compat=1.18}
\usetikzlibrary{arrows.meta, positioning, shapes.geometric, fit, calc, backgrounds, decorations.pathreplacing, matrix}

\title{OriginBlame: Record- and Token-Level Data Provenance for AI Training Datasets}
\author{Haolin Xue \\
  Northwestern Polytechnical University \\
  \texttt{hxue@mail.nwpu.edu.cn}}

\begin{document}

\maketitle

\begin{abstract}
	When a data contributor requests removal, model trainers face a practical gap: unlearning algorithms require a forget set, yet no tool can locate which training records belong to a given author. Existing provenance systems operate at file or dataset level, forcing catastrophic over-deletion. We present \texttt{ob}, a record- and token-level data provenance system that propagates author identity through data processing pipelines and resolves revocation requests into precise forget sets via deterministic queries. Evaluation on 219,555 Wikipedia pages demonstrates that record-level provenance eliminates dataset-level over-deletion (from 101$\times$ to 1.3$\times$), while integration adds 1.3--4.0\% throughput overhead (HuggingFace) and 2.1--19.0\% (Datatrove) on wiki data. On a 1.7B model, provenance-based forget sets consistently reduce the collateral damage of machine unlearning (retain perplexity) relative to same-size random baselines across all evaluated authors, with membership-inference tests indicating they select genuinely memorized content.
\end{abstract}

\section{Introduction}

\begin{figure}[t]
	\centering
	\includegraphics[width=\columnwidth]{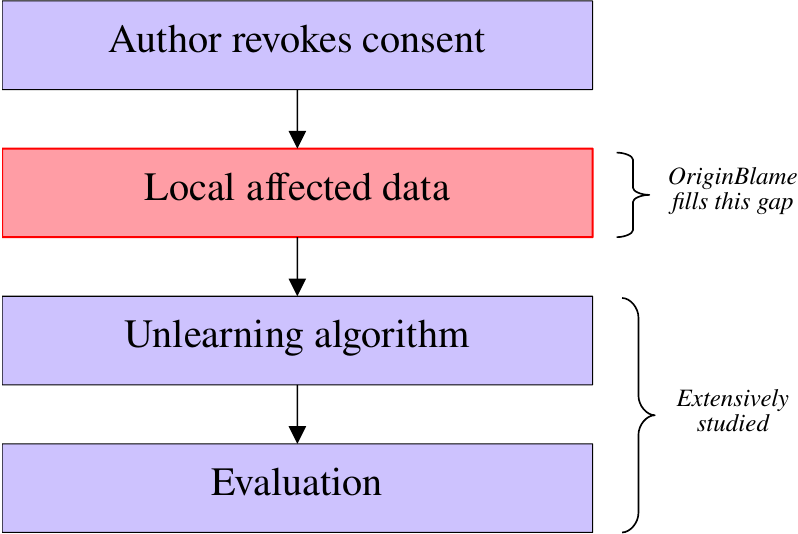}
	\caption{The unlearning pipeline gap. Existing methods assume a pre-defined forget set, but locating affected data remains unsolved. OriginBlame fills this gap.}
	\label{fig:unlearning-gap}
\end{figure}

When a data contributor requests removal, the model trainer must identify and remove that contributor's data from the training set. Machine unlearning methods such as Negative Preference Optimization (NPO)~\cite{zhang2024npo} and Representation Misdirection for Unlearning (RMU)~\cite{li2024wmdp} can degrade the influence of specific data once a forget set is provided, but all existing research assumes this forget set is already known. In practice, training datasets are assembled from thousands of sources through data processing, tokenization, and packing; by the time the dataset reaches the trainer, the connection between individual training lines and their original contributors has been lost. Without fine-grained provenance, the only options are catastrophic over-deletion (discard the entire dataset) or imprecise post-hoc inference~\cite{dangelo2025forsid}. Forget-set quality directly determines unlearning effectiveness. Benchmarks such as TOFU~\cite{maini2024tofu} and MUSE~\cite{shi2025muse} evaluate unlearning algorithms on synthetically curated forget sets, but their results cannot translate to real-world compliance without a mechanism to locate affected data at training granularity. Figure~\ref{fig:unlearning-gap} illustrates this gap between knowing \emph{who} to forget and knowing \emph{which data} to forget.

As shown in Table~\ref{tab:provenance-comparison}, data version control tools (DVC \cite{dvc}, LakeFS \cite{lakefs}, Delta Lake \cite{deltalake}) operate at file or dataset level. Experiment management tools (MLflow \cite{mlflow}, Weights \& Biases \cite{wandb}) track metadata but not data origins. Provenance tools like yProv4ML \cite{padovani2025yprov4ml} capture dataset-level relationships, while DLProv~\cite{pina2025dlprov} requires extensive instrumentation. None provide record-level provenance with author attribution.

OriginBlame (ob) tracks which training records derive from which source sections and which authors contributed to those sections, via a three-layer architecture (Figure~\ref{fig:three-layer}) with an independent token-index layer. Users integrate with one \texttt{ob.track()} call per record. ob provides precise forget sets at both record and token granularity.

We emphasize that ob does not discover authorship; it \emph{propagates} it. ob applies to data collection pipelines where author identity is already established by the source environment---for example, MediaWiki revision histories or GitHub commits---but this attribution is \emph{lost} during tokenization and packing. ob preserves the provenance chain across these transformations, enabling downstream queries (e.g., ``find all training lines from author~$a_i$'') that would otherwise require post-hoc inference.

\begin{table}[t]
	\centering
	\caption{Provenance Tools Comparison}
	\label{tab:provenance-comparison}
	\resizebox{\columnwidth}{!}{%
		\begin{tabular}{lcccc}
			\toprule
			\textbf{Tool/System}                   & \textbf{Granularity}       & \textbf{Author Revoke?} & \textbf{License?} & \textbf{Adoption (Stars)} \\
			\midrule
			DVC \cite{dvc}                         & File/dataset               & $\times$                & $\times$          & 15k+                      \\
			LakeFS \cite{lakefs}                   & Dataset/table              & $\times$                & $\times$          & 5k+                       \\
			Delta Lake \cite{deltalake}            & Table                      & $\times$                & $\times$          & 5k+                       \\
			MLflow \cite{mlflow}                   & Experiment                 & $\times$                & $\times$          & Very active               \\
			W\&B \cite{wandb}                      & Experiment                 & $\times$                & $\times$          & Very active               \\
			yProv4ML \cite{padovani2025yprov4ml}   & Dataset                    & $\times$                & $\times$          & Low (Jul 2025)            \\
			DLProv \cite{pina2025dlprov}           & Task                       & $\times$                & $\times$          & Low                       \\
			Dolt \cite{dolt}                       & Row (last modifier)        & $\times$                & $\times$          & 18k+                      \\
			WikiWho \cite{flock2014wikiwho}        & Token                      & $\times$                & $\times$          & Academic                  \\
			\rowcolor{blue!10}\textbf{OriginBlame} & \textbf{Line (authorship)} & \textbf{$\checkmark$}   & \textbf{Yes}      & \textbf{--}               \\
			\bottomrule
		\end{tabular}%
	}
\end{table}

This paper makes three contributions. First, we design a three-layer content-addressable architecture (authors $\leftarrow$ sections $\leftarrow$ document-index) with no ML or GPU dependencies. Second, we show that this architecture resolves author revocation requests into precise forget sets via deterministic queries and demonstrate that record-level provenance eliminates the dataset-level over-deletion that file-level tools incur. Third, we introduce an independent token-index layer that records source attribution during tokenization and packing, extending forget-set production to token granularity without requiring document-index records.

\section{Problem Formulation}

D'Angelo et al.~\cite{dangelo2025forsid} formalize the Forget-Set Identification (ForSId) problem: given a training set $D$, a model $M_D$, an unwanted set $D_u$, and a wanted set $D_w$, find a forget set $D_f \subseteq D$ whose removal maximally preserves model behavior on $D_w$ while altering it on $D_u$. ForSId requires model access to compute per-sample influence via training-data gradients. More fundamentally, its input is \emph{behavioral} evidence---samples whose predictions should change---not metadata-level queries such as ``forget all data by author~$a_i$.''

When author $a_i$ initiates a withdrawal request, the target forget set is $F = \{l \in D \mid \text{author}(l) = a_i\}$. Computing $F$ precisely without model access is the challenge ob addresses. Dataset-level provenance can only answer ``data from author~$a_i$ exists in~$D$,'' forcing two extremes: delete the entire dataset or take no action. Line-level provenance resolves this by answering ``author~$a_i$ contributed $n$ lines,'' enabling targeted deletion---as we show in Section~\ref{sec:revocation-precision}, record-level revocation reduces over-deletion by up to 101$\times$ compared to dataset-level approaches.

From this gap, we derive three design requirements. \emph{Precision}: locate specific data lines, not files or datasets. \emph{Verifiability}: provenance must be independently auditable, not merely stored internally. \emph{Minimal invasiveness}: integration must require only a few lines of code---DLProv~\cite{pina2025dlprov} serves as a cautionary example, requiring extensive script instrumentation despite low runtime overhead.

\section{Related Work}

As shown in Table~\ref{tab:provenance-comparison}, existing provenance tools stop at file or dataset granularity. Chen et al.~\cite{chen2026finegrained} propose fine-grained sample-level traceability with contribution scores across ML pipelines but focus on training-stage data usage auditing rather than authorship attribution. OriginBlame coexists with tools like DVC and HuggingFace Datasets: DVC manages file versions, while ob manages record-level author provenance.

Machine unlearning research similarly assumes the forget set is known (Table~\ref{tab:unlearning-benchmarks}). Benchmarks TOFU~\cite{maini2024tofu} and OpenUnlearning use pre-defined forget sets, while MUSE~\cite{shi2025muse} evaluates unlearning on real-world corpora with known forget/retain splits. Unlearning methods like NPO, RMU, GradAscent, and exact unlearning algorithms~\cite{muresanu2025exact} directly receive forget sets as input. ForSId~\cite{dangelo2025forsid} (\S2) is the only work addressing forget-set identification, but operates post-hoc with model access. OriginBlame captures origin information at recording time, requiring no model access. \begin{table}[t]
	\centering
	\caption{Machine Unlearning Benchmarks: The Forget-Set Location Gap}
	\label{tab:unlearning-benchmarks}
	\resizebox{\columnwidth}{!}{%
		\begin{tabular}{lccc}
			\toprule
			\textbf{Benchmark/Method}              & \textbf{Forget Set Source}                  & \textbf{Identifies Location?} \\
			\midrule
			TOFU \cite{maini2024tofu}              & Pre-defined (fictitious authors)            & $\times$                      \\
			MUSE \cite{shi2025muse}                & Pre-defined (2 corpora)                     & $\times$                      \\
			OpenUnlearning                         & Pre-defined                                 & $\times$                      \\
			NPO/RMU                                & Pre-defined                                 & $\times$                      \\
			ForSId \cite{dangelo2025forsid}        & Post-hoc inference                          & Partial                       \\
			\rowcolor{blue!10}\textbf{OriginBlame} & \textbf{Author revocation $\to$ forget set} & \textbf{$\checkmark$}         \\
			\bottomrule
		\end{tabular}%
	}
\end{table}

This positioning complements methods like Attribute-to-Delete~\cite{georgiev2024attribute}, which use datamodels to simulate unlearning but still require the forget set as input.

IPFS's content-addressing design~\cite{trautwein2022ipfs} inspired OriginBlame's hash chain mechanism, but IPFS operates at chunk level (256\,KB blocks) while OriginBlame refines granularity to lines. OriginBlame uses plain JSONL rather than a special file system. Compared to git-blame, designed for code editing scenarios, OriginBlame targets generated data scenarios: supporting multi-source attribution where a single line may fuse contributions from multiple authors, and author-initiated revocation.

\section{System Design}

\subsection{Architecture Overview}

OriginBlame (abbreviated as ob) stores all provenance metadata in a \texttt{.ob/} directory colocated with the project data. All files are plain JSONL with no configuration files or central database.

\begin{figure}[t]
	\centering
	\includegraphics[width=\columnwidth]{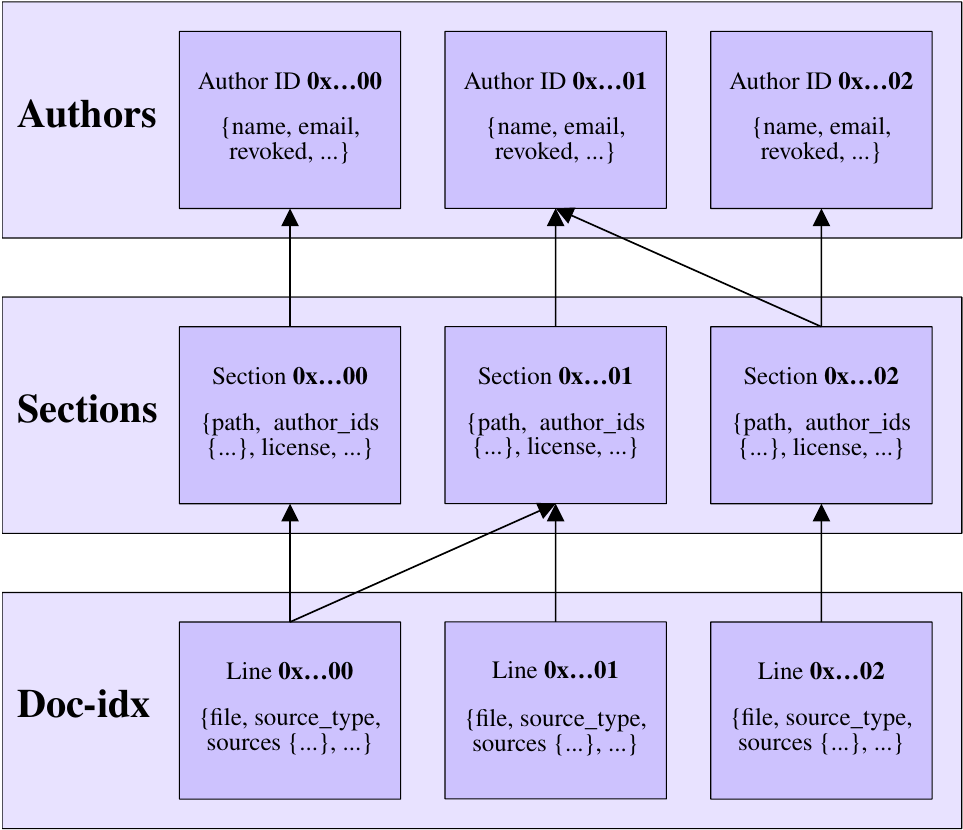}
	\caption{Three-layer reference architecture with hash-based sharding.}
	\label{fig:three-layer}
\end{figure}

\textbf{Three-tier model}. ob employs a hierarchical storage architecture. As illustrated in Figure~\ref{fig:three-layer}, three core layers are linked by content-addressable hashes in a strict parent--child relationship:

\textbf{Authors layer} (top) stores identities and revocation tags: \texttt{id} (SHA-256 of \texttt{name+email}), \texttt{name}, \texttt{email}, and \texttt{revoked} (boolean). The \texttt{revoked} field is the single source of truth for revocation, cascading lazily at query time; \texttt{email} serves as the lookup key for \texttt{revoke}. \textbf{Sections layer} (middle) stores file-level copyright: \texttt{section\_hash} (primary key, SHA-256 of \texttt{\{path, authors, license, year\}}), \texttt{path} (source file path, groups records by file), \texttt{authors} (list of author ids), an optional \texttt{contributors} list (page-level contributors recorded alongside primary authors, used where per-page attribution differs from section authorship), \texttt{license}, \texttt{year}, and \texttt{revoked} (boolean). \textbf{Document-index layer} (bottom) stores per-record provenance: \texttt{line\_hash} (SHA-256 of data content), \texttt{file} (output data file), \texttt{sources} (list of section hashes identifying contributing sources), \texttt{source\_type} (origin indicator, e.g., \texttt{"track"} for explicit tracking and \texttt{"reconcile"} for records relinked by reconcile), and \texttt{revoked} (boolean). The unique key is \texttt{(line\_hash, file, sources)}. Each document-index entry corresponds to one output record (one line in the data file), linked to its contributing sections and their authors. Document-index records do not store line numbers---\texttt{blame} computes hashes from data files at query time.

\textbf{Storage}. All core files are sharded into 256 buckets by the first two hexadecimal characters of their primary hash: \texttt{document-index/00}--\texttt{/ff}, \texttt{sections/00}--\texttt{/ff}, \texttt{authors/00}--\texttt{/ff}. Empty buckets are not created. This design enables O(1) lookup---given a \texttt{line\_hash}, \texttt{blame} directly reads \texttt{document-index/\{first two hex chars\}} without scanning unrelated buckets. Multi-process safety is achieved without locks: each process writes to isolated files (\texttt{docidx.\{pid\}}, \texttt{lock.\{pid\}}), which \texttt{ob clean} later merges into sharded buckets. Unmerged pid files block read operations until clean completes. An \texttt{embeddings.\{model\}/} directory stores per-model embedding vectors sharded into 256 buckets for reconcile (\S\ref{sec:package-ecosystem}); the core requires no ML libraries or GPU---embeddings are computed by the reconcile utility (\S\ref{sec:package-ecosystem}), not the user.

\textbf{Content-addressable hashing}. ob hashes dict inputs by serializing to JSON with sorted keys then applying SHA-256~\cite{nist2015fips1804}; str inputs are hashed as raw UTF-8 bytes. No normalization is performed. The collision probability for $n$ records over a $k$-bit hash space is bounded by the birthday approximation $p \approx n^2 / 2^{k+1}$~\cite[Thm.~A.4]{katz2020moderncrypto}. At the largest evaluated scale ($n = 2.2 \times 10^5$ records, $k = 256$), this yields $p < 10^{-68}$---negligible relative to any practical failure mode.

\textbf{Design principles}. ob follows two core principles. First, \emph{what is not recorded does not exist}: ob performs no post-hoc inference and does not mark unprovenanced content as provenanced. Second, \emph{single-hop provenance}: the system tracks only direct mappings from raw sources to final output, without tracing intermediate processing stages. Deleting raw data files does not affect provenance or revocation.

\subsection{Core Workflow}

OriginBlame is designed as both a library and a CLI tool. The core is a Rust native implementation (mmap-based reads, rayon parallelism, binary indexes); the \texttt{ob} Python package, installable from PyPI (\texttt{pip install originblame}), is the primary distribution channel and delegates performance-critical operations to the native backend through a three-tier fallback chain: PyO3 bindings compiled into the Rust library (fastest path, 38 bound operations), the standalone Rust \texttt{ob} binary invoked as a subprocess, and a pure-Python fallback for environments without a native build. The package ships prebuilt wheels for major platforms via CI, so no Rust toolchain is required for installation. Users initialize a project with \texttt{ob init}, creating the \texttt{.ob/} directory; the Python init wrapper additionally writes a \texttt{.gitignore} for temporary files. One-time setup requires three CLI calls and one Python API call: (1)~\texttt{ob init}, (2)~\texttt{ob author.add NAME EMAIL}, (3)~\texttt{ob register.add --path PATH --authors AUTHOR --license LICENSE --year YEAR}, (4)~\texttt{source.append(PATH)} (Python API). Integration into an existing data pipeline requires adding a single \texttt{track()} call at the point where each training record is written to disk---for example, inserting \texttt{track(record, file="data/train.jsonl")} immediately before writing the record to the output file. The user's existing pipeline logic remains unchanged; ob only observes the data passing through.

\textbf{Source management}: \texttt{source.append(path)} resolves paths to section records and activates them, registering all sections associated with that path. An optional \texttt{section} parameter selects a single section by its hash. \texttt{source.pop()} removes the most recently appended source, or a specific source by file path. \texttt{with ob.sources(...):} provides scoped tracking. Track operations associate with all active sources automatically---the caller does not pass sources explicitly.

\begin{figure}[t]
	\centering
	\includegraphics[width=\columnwidth]{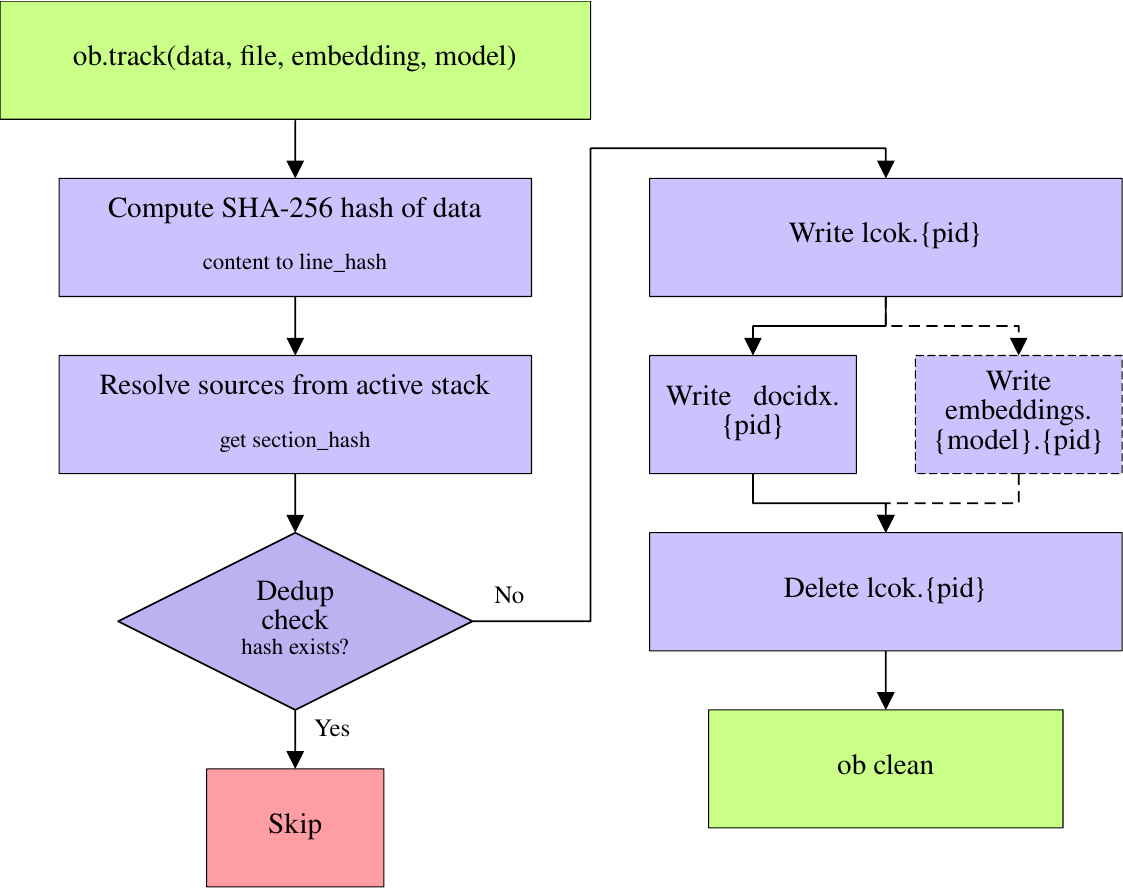}
	\caption{The track() workflow. After hash computation, the system checks for duplicates and writes to process-isolated temporary files under WAL protection.}
	\label{fig:track-flow}
\end{figure}

\textbf{ob.track() workflow}: The \texttt{track(data, file)} library function records provenance for a single data entry. Source attribution is resolved either from an explicit \texttt{source=} parameter (a file path that resolves to registered sections, or a list of section hashes) or, by default, from the source stack. As shown in Figure~\ref{fig:track-flow}, it (1)~computes the SHA-256 hash of \texttt{data} (JSON serialization for all types), (2)~resolves the active source list into section hashes, (3)~checks for duplicates (idempotent), and (4)~writes to process-isolated files via WAL (\texttt{lock.\{pid\}} $\rightarrow$ \texttt{docidx.\{pid\}}) in the Python API (the Rust native implementation writes directly to sharded buckets). If the process crashes before the lock is deleted, remaining data can be recovered by \texttt{ob clean}. An optional \texttt{embedding} parameter (Python API) stores a per-record embedding vector alongside the document-index record for later use by reconcile (\S\ref{sec:package-ecosystem}).

\textbf{Query functionality}: ob provides two orthogonal query methods. \textbf{ob blame} performs a forward query: given a file path and line number, it reads the line, computes SHA-256 to obtain \texttt{line\_hash}, queries \texttt{document-index/\{line\_hash[:2]\}}, and traverses the provenance chain: \texttt{line\_hash $\rightarrow$ sources (section\_hashes) $\rightarrow$ sections (author\_ids, license, path) $\rightarrow$ authors (name, email)}. \textbf{ob show} performs a reverse query: given \texttt{-{}-author}, it traverses \texttt{authors $\rightarrow$ sections $\rightarrow$ document-index} without reading data files. With \texttt{-{}-index}, both commands use the bucket-routing index (Appendix~\ref{sec:index}) for faster lookup. Additional filters include \texttt{-{}-section} (filter by section hash), \texttt{-{}-license} (filter by license name), \texttt{-{}-revoked} (show only revoked entries), and \texttt{-{}-tokenizer} (token-level granularity); these can be combined for intersection queries. \texttt{ob status} displays aggregate author counts and, with \texttt{-{}-tokenizer}, token-index statistics including entry counts, active/revoked breakdowns, and total token counts.

\textbf{Operation log}: The \texttt{.ob/log} file records state-changing operations (author/section registration, revocation, purge, clean, merge) as append-only log lines with timestamps, providing an audit trail for administrative actions. Read-only operations (blame, show, status) and per-record tracking are excluded to avoid volume overhead. \texttt{ob clean} rotates the log. \texttt{ob log} queries the log with optional \texttt{-{}-op} and \texttt{-{}-since} filters.

\begin{figure}[t]
	\centering
	\includegraphics[width=\columnwidth]{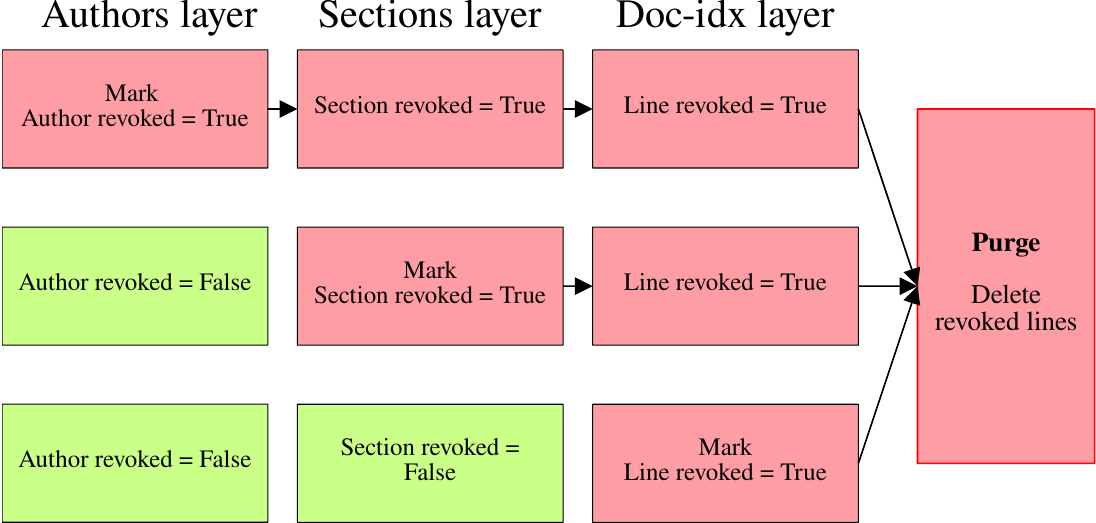}
	\caption{Revocation levels in ob. Author revoke cascades lazily from authors to sections to document-index. Section-level revoke marks sections. Purge physically deletes revoked entries after prior revoke.}
	\label{fig:revocation}
\end{figure}

\subsection{Revocation and Purge}

ob supports revocation at three granularity levels (Figure~\ref{fig:revocation}), all based on a tag model rather than bulk deletion. The single source of truth for revocation resides on the author record; all authors are equal---a contributor of a single line holds the same revocation rights as a contributor of the entire dataset.

\textbf{Author-level revoke}. \texttt{ob revoke -{}-author "z@e.com"} sets \texttt{revoked=true} on the author record. No other files are modified at this point. When a subsequent query (e.g., \texttt{blame} or \texttt{show}) encounters a revoked author, it evaluates the revocation status lazily along the chain: scan authors for \texttt{revoked=true} $\rightarrow$ find all sections whose \texttt{author\_ids} include that author $\rightarrow$ collect all document-index entries referencing those section hashes. The result is a \texttt{(file, line\_hash)} list of every data line traceable to the revoked author. The \texttt{-{}-reverse} flag undoes a revocation by toggling the tag back to \texttt{false}; with \texttt{-{}-tokenizer}, revocation also marks matching token-index entries.

\textbf{Section-level revoke}. \texttt{ob revoke -{}-section HASH} toggles \texttt{revoked=true} on a specific section record, affecting only the document-index entries that reference that section without revoking the author. This handles cases where a single file or data source must be withdrawn without affecting other content from the same contributor.

\textbf{Line-level revoke}. \texttt{ob revoke -{}-line-hash HASH -{}-file FILE} toggles \texttt{revoked=true} on a single document-index entry, enabling granular line-level withdrawal. This is the finest revocation granularity, useful when a specific data line must be removed without affecting any other lines from the same section or author.

\textbf{Purge}. \texttt{ob purge --file FILE} physically deletes revoked data lines from the specified file, leaving document-index cleanup to \texttt{ob clean} which archives revoked metadata into \texttt{archive/}. A safety constraint enforces \textbf{no purge without prior revoke}: purge can only delete lines that belong to revoked authors or revoked sections, preventing accidental data loss. The \texttt{-{}-index} variant routes directly to the relevant index buckets for faster operation. The \texttt{-{}-dry-run} flag previews which lines would be deleted without modifying files. The \texttt{-{}-reverse} flag restores previously purged lines from the archive directory.

\textbf{Lazy cascade rationale}. All revocation levels share a common design: the revoke operation modifies only a tag field on a single record. By default, queries exclude revoked entries (document-index and token-level alike); the \texttt{-{}-revoked} flag explicitly requests revoked entries. This yields two properties: (1)~revoke is lightweight regardless of how many data lines are affected; (2)~revocation is fully reversible---canceling a revoke removes the tag with immediate effect.

\subsection{Package Ecosystem}
\label{sec:package-ecosystem}

The ob architecture separates a core package from an optional utility package for advanced features.

\textbf{Core package} provides CLI commands for setup (\texttt{init}, \texttt{author.add}, \texttt{register.add}), source management via Python API (\texttt{source.append/pop}), querying (\texttt{blame}, \texttt{show}, \texttt{status}), audit logging (\texttt{log}), revocation (\texttt{revoke}, \texttt{purge}), and maintenance (\texttt{clean}, \texttt{merge}, \texttt{index build}, \texttt{generate-set}, \texttt{version}). The core compiles to a single native binary; users integrate via \texttt{author.add + register.add + source.append + track}.

\textbf{Utility features} (enabled via compile-time feature flags) provide optional parsers (\texttt{parse --parser mediawiki}), reconcile (\texttt{reconcile}), and export (\texttt{export-copyright}). The core binary performs zero imports of utility modules; when compiled without these features, ob omits the corresponding commands entirely.

\begin{figure}[t]
	\centering
	\includegraphics[width=\columnwidth]{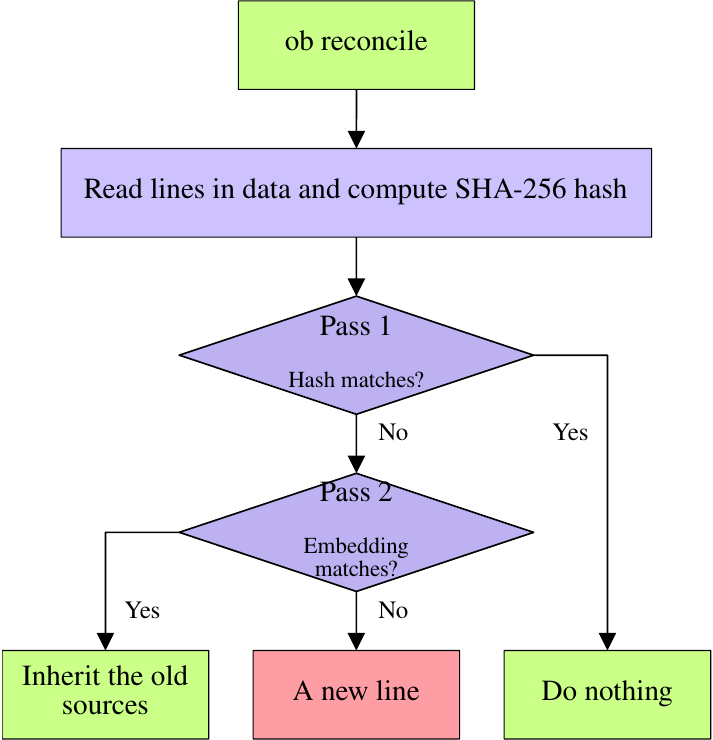}
	\caption{Two-phase reconcile strategy. Phase 1 attempts exact hash matching; Phase 2 falls back to embedding-based semantic matching. Unmatched lines are flagged as new.}
	\label{fig:reconcile}
\end{figure}

\textbf{Reconcile mechanism}: Reconcile operates in two phases (Figure~\ref{fig:reconcile}). Pass~1 attempts hash exact matching: each line is hashed and looked up in the document-index; a hit means unchanged content. Pass~2 performs embedding-based semantic matching via cosine similarity; a hit inherits the old record's sources, and unmatched old records are marked as orphans. Lines missing both phases require manual \texttt{track()}.

\subsection{Token-Level Provenance}

The three-layer architecture (authors $\leftarrow$ sections $\leftarrow$ document-index) captures per-record provenance, but tokenization and packing into binary shards eliminates record-level identity. ob addresses this with an independent \textbf{token-index} layer that links each document's token range in the packed binary to its source and author chain, operating in parallel to the document-index.

\textbf{Token-index entry}. Each entry stores \texttt{token\_count} (number of tokens produced from one document), \texttt{sources} (list of section hashes, linking to the author chain), \texttt{tokenizer} (identifier string, e.g., ``gpt2'' or ``llama3''), and \texttt{revoked} (boolean). The token-index is completely independent of the document-index: entries share the same source references but require no line hashes, and the two layers can exist with or without each other.

\textbf{Two operating modes}. In \emph{file mode}, the pipeline writes JSONL data files and calls \texttt{track()} to create document-index records; a subsequent tokenize+pack stage writes token-index entries. Both layers coexist, serving record-level and token-level queries respectively. In \emph{streaming mode}, the pipeline tokenizes documents directly without producing intermediate JSONL; only token-index entries are written, making the token-index the sole provenance record.

\textbf{Storage}. Token-index entries follow the same PID file pattern as document-index: each worker writes to \texttt{token-index.\{tokenizer\}.\{pid\}}, which \texttt{ob clean} merges into numbered files under \texttt{token-index.\{tokenizer\}/}. To preserve position-indexing correctness across parallel workers, \texttt{ob clean} merges PID files sequentially and computes cumulative token offsets only after all files are merged---not during write. Different tokenizers are stored in separate directories, following the same pattern as the optional embeddings directory. The tokenizer name is passed as a parameter; no tokenizer library is imported.

\textbf{Token-level queries}. Because entries are written sequentially during tokenization, the $i$-th entry's token range in the packed binary can be reconstructed from the cumulative sum of preceding entries' \texttt{token\_count} values, with its length given by its own \texttt{token\_count}. This makes the token-index a position-indexed provenance layer---no per-token records are needed. \texttt{ob show -{}-author NAME -{}-tokenizer gpt2} traverses author $\rightarrow$ section $\rightarrow$ token-index, summing token counts for all matching entries. \texttt{ob revoke -{}-author NAME -{}-tokenizer gpt2} marks matching entries as revoked. \texttt{ob generate-set -{}-tokenizer gpt2 -o forget.bin} produces a binary bitmask where each bit corresponds to one token-index entry (1 = revoked, 0 = active), directly usable by unlearning algorithms. \texttt{ob status -{}-tokenizer gpt2} reports aggregate token counts.

\textbf{Immutability constraint}. Because entry positions encode token ranges, the token-index does not support archive or purge operations. Revoked entries remain in the index with \texttt{revoked=true}; the bitmask produced by \texttt{generate-set} is the sole mechanism for conveying revocation to downstream unlearning algorithms. If the training pipeline is re-run with different data, the token-index is rebuilt from scratch rather than reconciled.

\textbf{Framework integration}. The integration pattern is framework-agnostic: intercept each data record, register provenance metadata, and pass the record through unmodified. We validate this with two frameworks. For Datatrove~\cite{penedo2024datatrove}, we implement OBProvenanceBlock as a PipelineStep whose \texttt{run()} method receives a generator of Documents, registers authors and sections from \texttt{Document.metadata}, writes token-index entries, and yields each document to the next stage---the same interface used by Datatrove's built-in blocks (e.g., LanguageFilter, TokensCounter). For HuggingFace Datasets, an equivalent integration wraps the \texttt{.map()} callback: the function loads zhwiki data via \texttt{load\_dataset()}, applies \texttt{author\_add + register\_section + track} per document, and writes provenance alongside output. Both integrations require the same $\sim$5-line adapter and produce identical \texttt{.ob/} directories; the downstream tokenizer requires no modification. Section~\ref{sec:token-eval} reports the overhead of both integrations.

\section{Evaluation}
\label{sec:evaluation}

This section validates that fine-grained provenance at both record and token granularity is necessary for practical author revocation. We measure revocation precision, reconcile recovery rates, storage overhead, query latency across four dataset scales, token-index integration overhead in a real processing framework, and whether provenance-based forget sets reduce the collateral damage of machine unlearning relative to random baselines. This is a functional validation on real data, not a comprehensive performance benchmark.

\subsection{Experimental Setup}
\label{sec:setup}

The evaluation dataset originates from a Chinese Wikipedia (zhwiki) dump comprising 219,555 wiki pages authored by 482,543 unique contributors under CC-BY-SA-4.0 licenses. The data construction pipeline has two stages.

\emph{Stage 1 --- Parse.} A stream parser extracts the latest-revision wikitext and the full contributor list from each page in the XML dump. Only namespace-0 articles with $\geq$50 characters and at least one contributor are retained, producing one source file per page and a metadata index (title, authors, year).

\emph{Stage 2 --- Track.} At each target scale, wiki source texts are written as JSONL records via HuggingFace Datasets or Datatrove pipelines, with \texttt{ob} provenance tracking inserted at the document processing stage. Each record receives author registration, section creation, and provenance linking through the \texttt{OBTrack} pipeline step. The resulting JSONL lines each track to their wiki page sources through the three-layer chain, while the Datatrove pipeline additionally produces token-index entries for token-level provenance queries.

\subsection{Revocation Precision}
\label{sec:revocation-precision}

The central question is whether record-level provenance granularity matters in practice. To answer it, we simulate author revocation at the 10k scale for four contributors with widely different contribution shares, using a HuggingFace tokenization pipeline on zhwiki data. Table~\ref{tab:revocation} reports the results.

\begin{table}[t]
	\centering
	\small
	\caption{Revocation precision at 10k scale (HuggingFace pipeline). Over-deletion measures how much more data dataset-level revocation destroys compared to ob's record-level precision.}
	\label{tab:revocation}
	\setlength{\tabcolsep}{4pt}
	\begin{tabular}{@{}lrrr@{}}
		\toprule
		Revoking Author    & Share  & ob (records) & Over-deletion \\
		\midrule
		InternetArchiveBot & 79.5\% & 7,953        & 1.3$\times$   \\
		Walter Grassroot   & 17.1\% & 1,712        & 5.8$\times$   \\
		KLBot2             & 5.0\%  & 499          & 20.0$\times$  \\
		HuangQQ            & 1.0\%  & 99           & 101.0$\times$ \\
		\midrule
		Dataset-level      & ---    & 10,000       & ---           \\
		\bottomrule
	\end{tabular}
\end{table}

Without record-level provenance, the only available action is to discard the entire training set---there is no intermediate file-level option, because wiki pages are collaborative documents that cannot be subdivided by author. The resulting over-deletion ranges from 1.3$\times$ (InternetArchiveBot, 79.5\% coverage) to 101$\times$ (HuangQQ, 1.0\%). Since each generated record derives from exactly one wiki page, file-level deletion is equivalent to record-level deletion here; remaining over-deletion comes from co-authorship within pages.

The over-deletion factor follows a heavy-tailed distribution across all authors in the dataset: the four benchmark authors in Table~\ref{tab:revocation} span two orders of magnitude in over-deletion. This pattern is consistent across all four tested scales (1k--220k): the largest-share author (InternetArchiveBot, 63--80\% coverage) consistently has 1.3--1.6$\times$ over-deletion, while the smallest-share authors ($\approx$1\%) consistently face $\approx$100$\times$ over-deletion under dataset-level revocation. All four authors have 100\% of their records in multi-author sections---meaning that even record-level revocation affects shared content. This co-deletion is inherent to collaborative authorship.

\subsection{Reconcile Effectiveness}
\label{sec:reconcile-effectiveness}

When data files are edited after initial tracking, provenance links break because content hashes no longer match. We evaluate the reconcile mechanism (\S\ref{sec:package-ecosystem}) with deterministic mutations (seed\,=\,42): 10\% text edited, 5\% deleted, 5\% new lines, using \texttt{nomic-embed-text-v1.5} at cosine threshold 0.85. Table~\ref{tab:reconcile} reports the two-phase recovery rates.

\begin{table}[b]
	\centering
	\footnotesize
	\caption{Reconcile recovery rates. Hash = Pass~1 exact. Semantic = Pass~2 embedding (nomic-embed-text-v1.5, cosine $\geq$ 0.85). Recovery = (hash + semantic) / mutated lines. Mutations: seed\,=\,42, 10\% text edited, 5\% deleted, 5\% new.}
	\label{tab:reconcile}
	\setlength{\tabcolsep}{4pt}
	\begin{tabular}{@{}lrrrrr@{}}
		\toprule
		Scale & Hash & Semantic & Recovered & New & Recovery \\
		\midrule
		1k    & 865   & 103   & 968   & 37  & 96.3\%   \\
		10k   & 8,479 & 1,294 & 9,773 & 192 & 98.1\%   \\
		100k  & 84,821 & 13,222 & 98,043 & 1,827 & 98.2\%   \\
		\bottomrule
	\end{tabular}
\end{table}

Hash matching recovers 85--87\% of original lines; embedding similarity recovers an additional 10--13\%, bringing total recovery to 96--98\% across all scales. Recovery improves with dataset size (96.3\% at 1k to 98.2\% at 100k). The remaining 2--4\% are genuinely new or heavily rewritten content requiring manual re-tracking via \texttt{ob.track()}.

\subsection{Scalability}
\label{sec:scalability}

Table~\ref{tab:scalability} reports operation latency across four scales from 1k to 220k lines, averaged over three runs per scale, using real zhwiki data from the HuggingFace pipeline. The implementation uses rayon for parallel bucket scanning and a binary index (Appendix~\ref{sec:index}) that stores author-to-section routing. \texttt{show} and \texttt{purge} use parallel rayon workers to scan 256 bucket files concurrently; at 220k lines, \texttt{show} completes in 80\,ms and \texttt{revoke} in 22\,ms.

\begin{table}[t]
	\centering
	\footnotesize
	\caption{Operation latency across scales (3-run avg., ms). HuggingFace pipeline data: 1k--220k lines. \emph{idx} = binary index with author-to-section routing.}
	\label{tab:scalability}
	\setlength{\tabcolsep}{3pt}
	\begin{tabular}{@{}lrrrrrr@{}}
		\toprule
		& \multicolumn{2}{c}{show} & revoke & purge & \multicolumn{2}{c}{purge\_author} \\
		\cmidrule(lr){2-3}\cmidrule(lr){4-4}\cmidrule(lr){5-5}\cmidrule(lr){6-7}
		Scale & base & idx & & dry & base & idx \\
		\midrule
		1k              & 4.6  & 4.5  & 1.6  & 0.005 & 1.4  & 1.4  \\
		10k             & 6.8  & 7.4  & 2.6  & 0.005 & 1.8  & 1.8  \\
		100k            & 38.6 & 38.8 & 11.5 & 0.008 & 9.3  & 9.3  \\
		220k            & 79.7 & 80.6 & 21.9 & 0.006 & 17.0 & 17.0 \\
		\bottomrule
	\end{tabular}
\end{table}

Line coverage is 100\% at all scales, confirming every data line has a complete provenance chain.

\subsection{Token-Level Provenance Evaluation}
\label{sec:token-eval}

We evaluate the token-index layer on three axes: integration overhead in a real processing framework, query performance, and metadata preservation.

\emph{Framework integration}. We validate the integration pattern with two frameworks on zhwiki data at four scales: from 1k pages (2.8M gpt2 tokens, 4.8k authors) to 220k pages (712M tokens, 483k authors). For Datatrove~\cite{penedo2024datatrove}, Pipeline~A is a vanilla pipeline (reader $\rightarrow$ tokenizer $\rightarrow$ packed output); Pipeline~B inserts OBProvenanceBlock between reader and tokenizer. For HuggingFace Datasets, an equivalent \texttt{.map()} callback applies provenance tracking per document. Table~\ref{tab:token-overhead} reports both frameworks across all four scales.

\begin{table}[b]
\centering
\caption{Token-level provenance: integration overhead across four scales on zhwiki data. Pipeline~A: vanilla. Pipeline~B: with provenance tracking. Query time is \texttt{ob show -{}-author -{}-tokenizer gpt2}.}
\label{tab:token-overhead}
\footnotesize
\setlength{\tabcolsep}{2.5pt}
\begin{tabular}{@{}rrrrrr@{}}
\toprule
& & \multicolumn{2}{c}{\textbf{Throughput drop}} & \textbf{Storage} & \textbf{Query} \\
\cmidrule(lr){3-4}
\textbf{Pages} & \textbf{Tokens} & Datatrove & HF & (Datatrove) & (ms) \\
\midrule
1\,000 & 2.8\,M & $-$13.8\% & $-$2.0\% & 1.33$\times$ & 3 \\
10\,000 & 25.9\,M & $-$19.0\% & $-$2.5\% & 1.29$\times$ & 9 \\
100\,000 & 302.4\,M & $-$13.4\% & $-$1.3\% & 1.24$\times$ & 33 \\
219\,555 & 712.4\,M & $-$2.1\% & $-$4.0\% & 1.23$\times$ & 69 \\
\bottomrule
\end{tabular}
\end{table}

At the 1k-page scale, the Datatrove provenance block adds 13.8\% throughput overhead while HuggingFace adds only 2.0\%, reflecting that provenance tracking is I/O-bound and HuggingFace's pipeline is already bottlenecked on tokenization. Storage overhead follows a consistent trend: from 1.33$\times$ at 1k pages to 1.23$\times$ at 220k pages, reflecting that structured JSONL provenance metadata grows sublinearly relative to token data. HuggingFace shows a 4.0\% throughput penalty at the 220k-page scale (712M tokens). The storage ratio is higher at small scales because \texttt{.ob/} stores structured metadata (author names, emails, section hashes, licenses) while the packed output is compact binary token IDs (2\,bytes each).

\emph{Query performance}. After Pipeline~B completes, \texttt{ob show -{}-author NAME -{}-tokenizer gpt2} traverses the full author $\rightarrow$ section $\rightarrow$ token-index chain in 3--69\,ms across the four tested scales (Table~\ref{tab:token-overhead}). At the 220k-page scale (712M tokens, 483k authors), \texttt{ob revoke} completes in 16\,ms.

\emph{Metadata preservation}. Pipeline~A produces packed binary output with document boundaries but \emph{no author information}---consistent with the behavior of Datatrove's DocumentTokenizer and Megatron-LM's preprocess\_data, which discard metadata during tokenization. Pipeline~B preserves the complete author $\rightarrow$ section $\rightarrow$ token-index chain, enabling provenance queries that are impossible in the vanilla pipeline.

\subsection{Cross-Domain Generalization}
\label{sec:cross-domain}

To evaluate whether ob generalizes beyond wiki-style text, we apply the same integration pattern to Linux kernel C source code. The attribution source is \texttt{git blame -e}, which provides line-level authorship for each \texttt{.c}/\texttt{.h} file---a more accurate attribution than the file-level \texttt{git log --name-only} approach used by The Stack~\cite{kocetkov2022thestack} and StarCoder~\cite{li2023starcoder}. Files are first filtered by commit frequency (top $N$ most-committed \texttt{.c}/\texttt{.h}/\texttt{.S}/\texttt{.rs}/\texttt{.dts}/\texttt{.dtsi} files), then attributed via \texttt{git blame} on the top 1,000 files (line-level, budget-capped), with bulk \texttt{git log} commit-frequency attribution for the remainder. We test three scales: 1\,000, 10\,000, and all $\approx$44k attributed source files (Table~\ref{tab:cross-domain}).

\begin{table}[t]
\centering
\caption{Linux kernel cross-domain benchmark. Attribution via \texttt{git blame -e} (line-level authorship on top $N$ \texttt{.c}/\texttt{.h} files). Over-deletion compares file-level against record-level precision (revoking \texttt{torvalds}).}
\label{tab:cross-domain}
\footnotesize
\setlength{\tabcolsep}{2pt}
\begin{tabular}{@{}rrrrrr@{}}
\toprule
& & \multicolumn{2}{c}{\textbf{Throughput drop}} & \textbf{Storage} & \textbf{Over-del.} \\
\cmidrule(lr){3-4}
\textbf{Files} & \textbf{Authors} & Datatrove & HF & (Datatrove) & \\
\midrule
1\,000 & 671 & $-$25.9\% & $-$0\% & 1.06$\times$ & 9$\times$ \\
10\,000 & 5\,285 & $-$41.2\% & $-$0\% & 1.02$\times$ & — \\
44\,222 & 6\,964 & — & $-$1\% & 1.01$\times$ & 1.3$\times$ \\
\bottomrule
\end{tabular}
\end{table}

Both frameworks integrate ob provenance tracking with no code changes to the pipeline pattern established in \S\ref{sec:token-eval}. Storage overhead stays under 1.06$\times$ across scales, and all ob queries complete under 10\,ms. The HuggingFace throughput penalty is negligible ($<$1\%); Datatrove shows higher overhead (25.9--41.2\%), driven by kernel source files being smaller and faster to process, which makes the provenance block a larger relative cost.

The over-deletion analysis reveals that file-level deletion remains wasteful even with accurate attribution: at the smallest scale (1\,000 files, 671 authors), revoking Linus Torvalds' contributions at file granularity would delete 9$\times$ more lines than necessary. At the full scale ($\approx$44k files), the over-deletion factor drops to 1.3$\times$.

\subsection{Application: Machine Unlearning}
\label{sec:mu}

This subsection asks whether that precision translates to a downstream task: given a revocation request, do ob's record-level forget sets enable machine unlearning with less collateral damage than page-level or same-size random baselines?

\emph{Setup.} We fine-tune Qwen3-1.7B~\cite{qwen3_2025} (full-parameter supervised fine-tuning, bf16, 1 epoch, lr\,=\,2$\times$10$^{-5}$, effective batch\,=\,16) on ChatML QA pairs generated from 10k zhwiki pages using a locally served Qwen3.5-9B. The SFT checkpoint serves as the shared starting point for all unlearning runs. We select three contributors spanning a range of contribution shares---InternetArchiveBot (35.9\% of fine-tuning tokens), Fozuxilai (11.3\%), and Ohtashinichiro (5.4\%)---and construct three forget-set conditions per author: \emph{line} uses ob's provenance to select exactly the records attributed to the target author, \emph{page} selects all records from co-authored pages (the page-level prototype that file-level tools would produce), and \emph{random} draws a same-size random subset from the full corpus (seed\,=\,42). The line and random sets contain the same number of records; the only difference is \emph{which} records are selected. Wiki contributors edit scattered, unrelated topics, making this a deliberately adversarial setting: individual contributions lack the semantic cohesion that would make targeted removal easy.

We apply two unlearning algorithms: NPO~\cite{zhang2024npo} ($\beta$\,=\,0.1, 5 epochs, lr\,=\,10$^{-5}$) and RMU~\cite{li2024wmdp} (mid-network layer 14, $\alpha$\,=\,50, steering coefficient 300, 400 steps, lr\,=\,5$\times$10$^{-5}$), for 18 unlearning runs plus the SFT baseline. Evaluation uses six metrics: forget-set perplexity~(PPL, higher\,=\,better forgetting), retain-set PPL~(lower\,=\,better utility preservation), forget-set ROUGE-L~(lower\,=\,less memorization), retain-set ROUGE-L~(higher\,=\,better preservation), membership-inference AUC on the forget set~(MIA), and extraction strength~(fraction of forget-set answers the model reconstructs from partial prompts; lower\,=\,less memorization).

\begin{table}[t]
	\centering
	\footnotesize
	\caption{Machine unlearning with ob line-level and page-level provenance vs.\ random baselines (Qwen3-1.7B, full-parameter fine-tuning, zhwiki QA data). Percentages are token share in the fine-tuning corpus. Arrows indicate desired direction. \textbf{Bold}: provenance set beats random on retain PPL within the same algorithm and author. MIA = membership-inference AUC on the forget set ($0.5$ = indistinguishable). Extraction = fraction of forget-set answers the model reconstructs from partial prompts.}
	\label{tab:unlearning}
	\setlength{\tabcolsep}{2pt}
	\begin{tabular}{@{}llrrrrrr@{}}
		\toprule
		& & \multicolumn{2}{c}{PPL} & \multicolumn{2}{c}{ROUGE-L} & MIA & Extr. \\
		\cmidrule(lr){3-4}\cmidrule(lr){5-6}
		Algo. & Set & forget $\uparrow$ & retain $\downarrow$ & forget $\downarrow$ & retain $\uparrow$ & $\to$0.5 & $\downarrow$ \\
		\midrule
		SFT & --- & 4.5 & 5.0 & 0.21 & 0.14 & 0.51 & 0.76 \\
		\midrule
		\multicolumn{8}{l}{\textit{InternetArchiveBot (35.9\%)}}\\
		NPO & line  & 8.3 & \textbf{7.0} & 0.09 & 0.10 & 0.77 & 0.50 \\
		NPO & page  & 8.3 & 7.5 & 0.07 & 0.15 & 0.61 & 0.64 \\
		NPO & rand  & 8.1 & 7.5 & 0.13 & 0.13 & 0.58 & 0.05 \\
		RMU & line  & 4.5 & 5.0 & 0.21 & 0.15 & 0.51 & 0.88 \\
		RMU & page  & 4.5 & 5.2 & 0.14 & 0.19 & 0.43 & 0.90 \\
		RMU & rand  & 4.8 & 4.8 & 0.19 & 0.20 & 0.53 & 0.92 \\
		\midrule
		\multicolumn{8}{l}{\textit{Fozuxilai (11.3\%)}}\\
		NPO & line  & 7.3 & \textbf{6.3} & 0.26 & 0.11 & 0.71 & 0.88 \\
		NPO & page  & 7.8 & \textbf{5.9} & 0.19 & 0.17 & 0.78 & 0.92 \\
		NPO & rand  & 8.2 & 7.5 & 0.07 & 0.10 & 0.47 & 0.45 \\
		RMU & line  & 3.5 & 5.1 & 0.29 & 0.12 & 0.27 & 0.84 \\
		RMU & page  & 3.0 & 5.1 & 0.28 & 0.18 & 0.23 & 0.82 \\
		RMU & rand  & 4.8 & 4.8 & 0.10 & 0.14 & 0.44 & 0.87 \\
		\midrule
		\multicolumn{8}{l}{\textit{Ohtashinichiro (5.4\%)}}\\
		NPO & line  & 7.3 & \textbf{6.5} & 0.09 & 0.10 & 0.67 & 0.79 \\
		NPO & page  & 7.2 & \textbf{5.5} & 0.11 & 0.15 & 0.83 & 0.96 \\
		NPO & rand  & 8.3 & 7.4 & 0.12 & 0.09 & 0.54 & 0.15 \\
		RMU & line  & 3.9 & 4.8 & 0.18 & 0.16 & 0.42 & 0.74 \\
		RMU & page  & 2.9 & 4.8 & 0.29 & 0.14 & 0.24 & 0.80 \\
		RMU & rand  & 4.9 & 4.7 & 0.18 & 0.16 & 0.52 & 0.73 \\
		\bottomrule
	\end{tabular}
\end{table}

\emph{The cost of unlearning.} Retain PPL measures how well the model still fits the content it should keep; every unlearning run in Table~\ref{tab:unlearning} sits above the SFT checkpoint (5.0), i.e., none leaves the retain distribution untouched. This cost is inherent to approximate unlearning: gradient updates against the forget set are applied to parameters shared by all content. The SFT row is not itself an option under a compliance request---the question this evaluation answers is not whether the cost occurs, but how forget-set quality determines how much of it each option pays. RMU illustrates the opposite end of the trade-off: it holds retain PPL at 4.7--5.2 only by achieving little forgetting (forget PPL 2.9--4.9 vs.\ 4.5 for SFT).

\emph{NPO results.} Provenance-based forget sets achieve comparable forgetting at consistently lower collateral damage (Table~\ref{tab:unlearning}). Forget PPL rises well above the SFT baseline (4.5) for all set types (7.2--8.3), but retain PPL is lower for line-level sets (6.3--7.0) than for random sets (7.4--7.5) in all three authors (7--16\%), with page-level sets improving further on two (up to 5.5 vs.\ 7.4 for Ohtashinichiro, 26\%). The MIA column is consistent with the mechanism: provenance sets score 0.61--0.83 AUC versus 0.47--0.58 for random, indicating content the model has distinctly memorized, whereas random sets dilute the unlearning signal with content the model barely learned. The mechanism is intuitive: a random set samples the corpus uniformly, so gradients against it generalize to the retain distribution, while provenance concentrates them on targeted content. This advantage is specific to the probability metric: retain ROUGE-L shows no consistent line-vs-random separation (0.10--0.11 vs.\ 0.09--0.13 across the three authors; differences within 0.03 with mixed sign).

\emph{RMU results.} RMU at these hyperparameters produces weak forgetting regardless of forget-set type: all metrics remain near the SFT baseline (forget PPL 2.9--4.9 vs.\ 4.5), and extraction stays high (0.73--0.92) across line, page, and random sets. The provenance advantage that emerges under NPO does not appear under RMU on this corpus; steering hidden states away from forget content appears to require per-corpus tuning beyond our fixed budget, a limitation independent of forget-set quality.

\emph{Transformation gap.} Even under NPO with provenance sets, extraction remains high (0.50--0.88). MIA indicates these sets contain genuinely memorized content, so persistent extraction is consistent with knowledge having been absorbed beyond record boundaries---though insufficient unlearning strength at our fixed hyperparameters cannot be excluded either. Wikipedia is the hardest case: a single contributor's edits are scattered across unrelated topics, and frequently occurring knowledge is shared across multiple contributors, making it harder to suppress~\cite{krishnan2025notall}. Per-record influence on model parameters is non-localized~\cite{hammoudeh2024influence}; bridging record-level provenance with knowledge-level unlearning remains an open challenge.

\emph{Takeaway.} Record-level provenance consistently reduces the collateral damage of unlearning relative to same-size random baselines at matched forgetting strength---five of six pairwise comparisons favor provenance on retain PPL, one ties---and membership-inference results are consistent with provenance selecting genuinely memorized content. Three caveats bound this claim: these are single-run results (seed 42) without variance estimates; each condition's retain metrics are computed on the complement of its own forget set, so retain scores are evaluated on overlapping but distinct line sets; and the retain-PPL advantage does not replicate on retain ROUGE-L, where line and random degrade comparably. We therefore emphasize the direction of the effect, consistent across all three authors on the probability metric, over its magnitude. The transformation gap adds a final bound: pointing at the source is necessary, not sufficient---precise forget sets must be paired with unlearning methods capable of erasing absorbed knowledge.

\section{Discussion}
\label{sec:discussion}

OriginBlame has five limitations. (1)~Incremental adoption is difficult: users cannot retroactively add provenance to existing datasets. (2)~The parser ecosystem is immature: only a MediaWiki parser is currently available, though our cross-domain evaluation (\S\ref{sec:cross-domain}) confirms that the core provenance tracking is domain-agnostic and works with arbitrary attribution sources like git blame. (3)~Single-hop provenance: the system tracks only direct mappings from raw data to final output. (4)~The index must be rebuilt when data changes; indexed variants do not consistently outperform the full-scan path due to rayon parallelism amortizing the scan cost. (5)~The unlearning evaluation is single-run (seed 42) on one model, one corpus, and three authors; the provenance advantage is established on retain PPL direction only and does not replicate on retain ROUGE-L (\S\ref{sec:mu}).

Future work will pursue three directions: expanding the parser ecosystem to support Common Crawl archives, PDF documents, and mainstream NLP dataset formats; exploring incremental adoption paths that allow users to partially embed OriginBlame in existing pipelines; and bridging the transformation gap---the unlearning results show that record-level provenance selects genuinely memorized content, yet extraction persists because knowledge is absorbed beyond record boundaries, so effective erasure requires coupling provenance with knowledge-level unlearning methods rather than record-level targeting alone.

\paragraph{Privacy and dual-use.}
The authors layer stores contributor names and emails alongside their provenance records. In production deployments handling sensitive contributor data, operators can strip PII from \texttt{.ob/} by removing name and email fields from author records---leaving only the SHA-256 identifier---and maintaining a separate encrypted mapping outside the provenance directory. The operation log (\texttt{.ob/log}) provides an audit trail for revocation and query commands, supporting access-control policies that restrict provenance queries to authorized compliance personnel. We note that fine-grained author attribution could be misused to target individual contributors; deploying organizations should restrict provenance query access accordingly.

\section{Conclusion}

Machine unlearning research has overlooked a prerequisite: locating the affected data. OriginBlame addresses this gap with a three-layer content-addressable architecture that propagates author identity through data processing pipelines, resolving revocation requests into precise forget sets without model access or post-hoc inference.

Evaluation across multiple domains and scales confirms practical deployability: record-level provenance eliminates the 1--2 orders-of-magnitude over-deletion inherent in dataset-level approaches; integration overhead remains modest across frameworks; and two-phase reconcile recovers 96--98\% of provenance links after mutations. On the downstream unlearning task, provenance-based forget sets consistently cut collateral damage relative to same-size random baselines while achieving comparable forgetting, consistent with forget-set quality directly affecting unlearning outcomes---though the remaining transformation gap shows that precise selection must be paired with erasure methods capable of removing absorbed knowledge.

The source code and evaluation scripts are available at \url{https://github.com/tzbkk/originblame}; the Rust implementation at \url{https://github.com/tzbkk/rust-originblame}.

\section*{Acknowledgements}

The author used OpenCode (GLM-5.1) for language editing and code assistance, and takes responsibility for all AI-assisted content.

\bibliography{references}

@inproceedings{shi2025muse,
  title     = {{MUSE}: Machine Unlearning Six-Way Evaluation for Language Models},
  author    = {Shi, Weijia and Lee, Jaechan and Huang, Yangsibo and Malladi, Sadhika and Zhao, Jieyu and Holtzman, Ari and Liu, Daogao and Zettlemoyer, Luke and Smith, Noah A. and Zhang, Chiyuan},
  booktitle = {Proceedings of the International Conference on Learning Representations (ICLR)},
  year      = {2025},
  url       = {https://arxiv.org/abs/2407.06460}
}

@inproceedings{maini2024tofu,
  title     = {{TOFU}: A Task of Fictitious Unlearning for {LLMs}},
  author    = {Maini, Pratyush and Feng, Zhili and Schwarzschild, Avi and Lipton, Zachary C. and Kolter, J. Zico},
  booktitle = {Proceedings of the Conference on Language Modeling (COLM)},
  year      = {2024},
  url       = {https://arxiv.org/abs/2401.06121}
}

@inproceedings{muresanu2025exact,
  title     = {Fast Exact Unlearning for In-Context Learning Data for {LLMs}},
  author    = {Muresanu, Andrei I. and Thudi, Anvith and Zhang, Michael R. and Papernot, Nicolas},
  booktitle = {Proceedings of the International Conference on Machine Learning (ICML)},
  series    = {Proceedings of Machine Learning Research},
  volume    = {267},
  pages     = {45272--45288},
  year      = {2025},
  url       = {https://arxiv.org/abs/2402.00751}
}

@article{georgiev2024attribute,
  title     = {Attribute-to-Delete: Machine Unlearning via Datamodel Matching},
  author    = {Georgiev, Kristian and Rinberg, Roy and Park, Sung Min and Garg, Shivam and Ilyas, Andrew and Madry, Aleksander and Neel, Seth},
  journal   = {arXiv preprint arXiv:2410.23232},
  year      = {2024},
  url       = {https://arxiv.org/abs/2410.23232}
}

@article{dangelo2025forsid,
  title     = {The forget-set identification problem},
  author    = {D'Angelo, Andrea and Gullo, Francesco and Stilo, Giovanni},
  journal   = {Machine Learning},
  volume    = {114},
  number    = {247},
  year      = {2025},
  publisher = {Springer},
  doi       = {10.1007/s10994-025-06897-9}
}

@article{padovani2025yprov4ml,
  title     = {{yProv4ML}: Effortless Provenance Tracking for Machine Learning Systems},
  author    = {Padovani, G. and Anantharaj, V. and Fiore, S.},
  journal   = {arXiv preprint arXiv:2507.01078},
  year      = {2025},
  url       = {https://arxiv.org/abs/2507.01078}
}

@inproceedings{chen2026finegrained,
  title     = {Fine-Grained Traceability for Transparent {ML} Pipelines},
  author    = {Chen, Liping and Liu, Mujie and Fayek, Haytham},
  booktitle = {Proceedings of The Web Conference (WWW)},
  year      = {2026},
  note      = {Accepted},
  url       = {https://arxiv.org/abs/2601.14971}
}

@article{pina2025dlprov,
  title     = {{DLProv}: a suite of provenance services for deep learning workflow analyses},
  author    = {Pina, D{\'e}bora and Kunstmann, Liliane and Chapman, Adriane and de Oliveira, Daniel and Mattoso, Marta},
  journal   = {PeerJ Computational Science},
  year      = {2025},
  doi       = {10.7717/peerj-cs.2985}
}

@inproceedings{trautwein2022ipfs,
  title     = {Design and Evaluation of {IPFS}: A Storage Layer for the Decentralized Web},
  author    = {Trautwein, Dennis and Raman, Aravindh and Tyson, Gareth and Castro, Ignacio and Scott, Will and Schubotz, Moritz and Gipp, Bela and Psaras, Yiannis},
  booktitle = {Proceedings of the ACM SIGCOMM Conference},
  year      = {2022},
  doi       = {10.1145/3544216.3544232}
}

@misc{dvc,
  title        = {{DVC}: Data Version Control},
  author       = {{Treeverse}},
  howpublished = {\url{https://dvc.org}},
  year         = {2024}
}

@misc{mlflow,
  title        = {{MLflow}: A Platform for the Machine Learning Lifecycle},
  author       = {{Databricks}},
  howpublished = {\url{https://mlflow.org}},
  year         = {2024}
}

@misc{wandb,
  title        = {Weights \& Biases},
  author       = {{Weights \& Biases}},
  howpublished = {\url{https://wandb.ai}},
  year         = {2024}
}

@misc{lakefs,
  title        = {{LakeFS}: Data Version Control for Data Lakes},
  author       = {{Treeverse}},
  howpublished = {\url{https://lakefs.io}},
  year         = {2024}
}

@misc{deltalake,
  title        = {{Delta Lake}: Open-Source Storage Layer for Data Lakes},
  author       = {{The Linux Foundation}},
  howpublished = {\url{https://delta.io}},
  year         = {2024}
}

@misc{dolt,
  title        = {{Dolt}: Version Controlled Database},
  author       = {{Dolthub}},
  howpublished = {\url{https://github.com/dolthub/dolt}},
  year         = {2024}
}

@inproceedings{flock2014wikiwho,
  title     = {{WikiWho}: Precise and Efficient Attribution of Authorship of Revisioned Content},
  author    = {Fl{\"o}ck, Fabian and Acosta, Maribel},
  booktitle = {Proceedings of the 23rd International Conference on World Wide Web (WWW)},
  pages     = {843--854},
  year      = {2014},
  doi       = {10.1145/2566486.2568026}
}

@article{zhang2024npo,
  title     = {Negative Preference Optimization: From Catastrophic Collapse to Effective Unlearning},
  author    = {Zhang, Ruiqi and Lin, Licong and Bai, Yu and Mei, Song},
  journal   = {arXiv preprint arXiv:2404.05868},
  year      = {2024},
  url       = {https://arxiv.org/abs/2404.05868}
}

@inproceedings{li2024wmdp,
  title     = {The {WMDP} Benchmark: Measuring and Reducing Malicious Use with Unlearning},
  author    = {Li, Nathaniel and Pan, Alexander and Gopal, Anjali and Yue, Summer and Berrios, Daniel and Gatti, Alice and Li, Justin D. and Dombrowski, Ann-Kathrin and Goel, Shashwat and Mukobi, Gabriel and Helm-Burger, Nathan and Lababidi, Rassin and Justen, Lennart and Liu, Andrew Bo and Chen, Michael and Barrass, Isabelle and Zhang, Oliver and Zhu, Xiaoyuan and Tamirisa, Rishub and Bharathi, Bhrugu and Herbert-Voss, Ariel and Breuer, Cort B. and Zou, Andy and Mazeika, Mantas and Wang, Zifan and Oswal, Palash and Lin, Weiran and Hunt, Adam Alfred and Tienken-Harder, Justin and Shih, Kevin Y. and Talley, Kemper and Guan, John and Steneker, Ian and Campbell, David and Jokubaitis, Brad and Basart, Steven and Fitz, Stephen and Kumaraguru, Ponnurangam and Karmakar, Kallol Krishna and Tupakula, Uday and Varadharajan, Vijay and Shoshitaishvili, Yan and Ba, Jimmy and Esvelt, Kevin M. and Wang, Alexandr and Hendrycks, Dan},
  booktitle = {Proceedings of the 41st International Conference on Machine Learning},
  series    = {Proceedings of Machine Learning Research},
  volume    = {235},
  pages     = {28525--28550},
  year      = {2024},
  url       = {https://proceedings.mlr.press/v235/li24bc.html}
}

@misc{penedo2024datatrove,
  author    = {Penedo, Guilherme and Kydl\'{i}\v{c}ek, Hynek and Cappelli, Alessandro and Sasko, Mario and Wolf, Thomas},
  title     = {{DataTrove}: Large Scale Data Processing},
  year      = {2024},
  publisher = {GitHub},
  journal   = {GitHub repository},
  url       = {https://github.com/huggingface/datatrove}
}

@article{kocetkov2022thestack,
  title     = {The Stack: 3 {TB} of permissively licensed source code},
  author    = {Kocetkov, Denis and Li, Raymond and Jivalds, Leandro and Jablonski, Thomas and de la Rosa, Marc and Li, Elizaveta and et al.},
  journal   = {Transactions on Machine Learning Research},
  year      = {2022},
  url       = {https://arxiv.org/abs/2211.15533}
}

@article{li2023starcoder,
  title     = {StarCoder: May the Source Be with You!},
  author    = {Li, Raymond and Li, Loubna Ben and Weber, Leandro and Xia, Meng and del-Rio, Manuel and Nie, Shang and et al.},
  journal   = {Transactions on Machine Learning Research},
  year      = {2023},
  url       = {https://arxiv.org/abs/2305.06161}
}

@book{katz2020moderncrypto,
  title     = {Introduction to Modern Cryptography},
  author    = {Katz, Jonathan and Lindell, Yehuda},
  edition   = {3rd},
  publisher = {CRC Press},
  year      = {2020},
  isbn      = {978-0-367-47562-2}
}

@misc{nist2015fips1804,
  title        = {{FIPS} 180-4: Secure Hash Standard ({SHS})},
  author       = {{National Institute of Standards and Technology}},
  howpublished = {\url{https://doi.org/10.6028/NIST.FIPS.180-4}},
  year         = {2015}
}

@misc{qwen3_2025,
  title         = {Qwen3 Technical Report},
  author        = {{Qwen Team}},
  howpublished  = {\url{https://qwenlm.github.io/blog/qwen3/}},
  year          = {2025}
}

@inproceedings{krishnan2025notall,
  title     = {Not All Data Are Unlearned Equally},
  author    = {Krishnan, Aravind and Reddy, Siva and Mosbach, Marius},
  booktitle = {Proceedings of the Conference on Language Modeling (COLM)},
  year      = {2025},
  url       = {https://arxiv.org/abs/2504.05058}
}

@article{hammoudeh2024influence,
  title     = {Training Data Influence Analysis and Estimation: A Survey},
  author    = {Hammoudeh, Zayd and Lowd, Daniel},
  journal   = {Machine Learning},
  year      = {2024},
  publisher = {Springer},
  doi       = {10.1007/s10994-023-06495-7},
  url       = {https://arxiv.org/abs/2212.04612}
}

\appendix

\section{Bucket-Routing Index}
\label{sec:index}

\texttt{show} and \texttt{purge} grow super-linearly because they scan all 256 document-index shard files and expand each record's section and author links. The bucket-routing index eliminates this full scan by pre-computing which shard files are relevant for each author and section.

\paragraph{Structure.}
The index is a fourth layer stored in \texttt{.ob/index/}, sharded by the same 2-character hex prefix as the other layers. Each shard file contains JSONL records of the form \texttt{\{"id": "<hash>", "refs": ["ab", "cd", ...]\}}, where \texttt{id} is either an author ID or a section hash, and \texttt{refs} lists the 2-character bucket prefixes of the next layer's relevant shards. Specifically:
\begin{itemize}
	\item \textbf{Author entries}: \texttt{id} = author ID, \texttt{refs} = section bucket prefixes containing that author.
	\item \textbf{Section entries}: \texttt{id} = section hash, \texttt{refs} = document-index bucket prefixes containing records that reference that section.
\end{itemize}
The index is built by a single pass over the document-index and section layers via \texttt{ob index build}.

\paragraph{Query path.}
For \texttt{show\,--\,author --index}, the query proceeds in three hops, each reading only the relevant shard files:
\begin{enumerate}
	\item Look up author entries in the index $\to$ obtain section bucket prefixes.
	\item Read those section shards to find matching sections $\to$ look up section entries in the index $\to$ obtain document-index bucket prefixes.
	\item Read only those document-index shards to collect results.
\end{enumerate}
\texttt{purge\,--\,author --index} follows the same three-hop path but returns only \texttt{(line\_hash, file)} pairs without expanding section and author details.

\paragraph{Implementation details.}
Two optimizations reduce I/O:
\begin{enumerate}
	\item \textbf{Batched bucket reads.} At each hop, IDs are grouped by their 2-character prefix so each shard file is read at most once, even when hundreds of IDs map to the same bucket.
	\item \textbf{Batch pre-loading.} For \texttt{show}, all author records referenced by matched sections are loaded into memory in a single pass over the author shards before the document-index loop, replacing per-result \texttt{get\_author()} calls with constant-time dict lookups.
\end{enumerate}

\paragraph{Unified binary index.}
The binary index (OBIDXF02) uses type-tagged references to serve both the document-index and token-index layers. Each reference has a 1-byte type tag: \texttt{0x00} for document-index shard prefixes and \texttt{0x01} for token-index byte ranges. A single section hash can reference both document-index shards and token-index ranges across multiple tokenizers. Queries filter by type and tokenizer to access only the relevant layer. In addition to OBIDXF02, token-index queries use a dedicated per-tokenizer binary index (OBIDXTI01, stored as \texttt{token-index.\{tokenizer\}.bin}) that maps sections to byte ranges within the merged token-index files.

\paragraph{Performance.}
Line-level query latency across all scales is reported in Table~\ref{tab:scalability} (\S\ref{sec:scalability}); token-index query latency in Table~\ref{tab:token-overhead} (\S\ref{sec:token-eval}). The index is opt-in via the \texttt{--index} flag; without it, both commands fall back to the parallel full-scan path.

\section{Reconcile Mutation Algorithm}
\label{sec:mutation}

The reconcile benchmark (\S\ref{sec:reconcile-effectiveness}) simulates data edits by applying three operations to the tracked data file with a deterministic random seed (42):

\begin{enumerate}
	\item \textbf{Edit} (10\%): Each line is independently classified by drawing $r \sim U(0,1)$. Lines with $r < 0.10$ are edited by replacing 10\% of their alphabetic characters with random lowercase letters. For JSON records, only the \texttt{text} field is modified; other fields (metadata, section hashes) are preserved.
	\item \textbf{Delete} (5\%): Lines with $0.10 \leq r < 0.15$ are removed entirely.
	\item \textbf{Insert} (5\%): $0.05 \times N$ new lines are generated by copying random existing lines and modifying 15\% of their alphabetic characters. Insertions are placed at random positions.
\end{enumerate}

This design preserves the character-set distribution of the original data (unlike random byte-flipping, which would produce invalid UTF-8 in Chinese text) while ensuring that edited and inserted lines have no hash overlap with original document-index entries, isolating the embedding-based semantic matching phase.

\end{document}